# Prediction of Unmanned Surface Vessel Motion Attitude Based on CEEMDAN-PSO-SVM

Zhuoya Geng，Jianmei Chen

**Abstract**：Unmanned boats, while navigating at sea, utilize active compensation systems to mitigate wave disturbances experienced by onboard instruments and equipment. However, there exists a lag in the measurement of unmanned boat attitudes, thus introducing unmanned boat motion attitude prediction to compensate for the lag in the signal acquisition process. This paper, based on the basic principles of waves, derives the disturbance patterns of waves on unmanned boats from the wave energy spectrum. Through simulation analysis of unmanned boat motion attitudes, motion attitude data is obtained, providing experimental data for subsequent work. A combined prediction model based on Complete Ensemble Empirical Mode Decomposition with Adaptive Noise (CEEMDAN), Particle Swarm Optimization (PSO), and Support Vector Machine (SVM) is designed to predict the motion attitude of unmanned boats. Simulation results validate its superior prediction accuracy compared to traditional prediction models. For example, in terms of mean absolute error, it improves by 17% compared to the EMD-PSO-SVM model.

With the development of society, intelligent unmanned platform technology has been rapidly improved, and unmanned boats, as representatives of intelligent unmanned platforms and marine technology equipment, have developed rapidly[1]. Unmanned boats have the advantages of small size, flexible use, fast speed, and strong adaptability. They have become an important engineering equipment for building a maritime power. They have broad application prospects in both civil and military fields, such as carrying sensors to detect marine resources, carrying professional equipment for marine scientific surveys, loading weapons and equipment to attack targets, etc.[2,3]. Therefore, unmanned boats typically operate in the complex and changeable marine environment, where they undergo six degrees of freedom motion, including heave, sway, surge, roll, pitch, and yaw, influenced by wind and waves. Especially in adverse sea conditions, unmanned boats sway rapidly, resulting in ineffective compensation of boat attitude by the stable platform and inability to maintain the stability of various instruments and equipment onboard, thus hindering normal maritime measurement and aiming tasks[4,5]. To address the issue of delayed compensation for platform attitude stabilization, an attitude prediction method for unmanned boats is proposed to anticipate boat attitude and compensate for errors introduced by various lagging factors. Therefore, studying the motion attitude of unmanned boats and predicting it holds profound significance.

Ship attitude prediction models can be classified into four types: models based on ship hydrodynamic motion equations, models based on time series data analysis, artificial intelligence models, and hybrid models [6]. In the 1960s, Kaplan [7] proposed using hydrodynamic theory to derive ship response kernel functions from ship motion response functions, convolving them with bow wave heights to achieve ship attitude prediction. However, due to difficulties in accurately obtaining bow wave heights and ship response kernel functions, this method failed to be widely applied in practical situations. With the development of modern control theory, Trantafyllou [8-10] proposed a ship attitude prediction method based on Kalman filtering. However, the forecasting results and stability were not satisfactory, especially in adverse sea conditions, failing to meet the requirements of model accuracy and noise statistics.

Methods based on time series data utilize past ship data to establish prediction models. This approach includes autoregressive (AR) models and autoregressive moving average (ARMA) models, which are computationally simple and have practical value[11,12]. However, AR and ARMA methods are not suitable for predicting nonlinear and non-stationary motion in adverse sea conditions [13].

To address the prediction of nonlinear ship motion attitudes in complex sea conditions, various artificial intelligence models based on machine learning have been proposed. Khan et al.[14] roposed a ship attitude prediction method based on artificial neural networks, which combines singular value decomposition and conjugate gradient algorithm, capable of predicting ship attitudes for up to 7 seconds. Yang et al. [15] introduced a ship attitude prediction model based on BP neural networks, characterized by high computational accuracy and real-time capability, providing rapid ship attitude prediction results, albeit with a high number of learning iterations. Yin [16,17] presented a ship attitude prediction model based on radial basis function networks and conducted simulation studies on ship roll motion prediction, demonstrating that the proposed neural network prediction model can accurately predict roll angles online. Wang et al. [18] established a ship attitude prediction model based on long short-term memory (LSTM) units and compared it with traditional AR models, showing that the model has advantages in accuracy and adaptability, suitable for predicting motion in various degrees of freedom of ships. Gu et al. [19] developed a ship motion attitude prediction model based on gated recurrent unit (GRU) neural networks, successfully achieving long-term, high-precision prediction of ship pitch and roll using short-term input data. Sun [20] proposed a ship pitch sequence prediction model based on support vector machines (SVM), introducing an RBF kernel function. This method effectively addresses the issues of small sample size and nonlinearity, demonstrating good generalization and prediction accuracy. In the process of using artificial intelligence models, the setting of parameters affects the accuracy of prediction. Therefore, parameter optimization of the model is crucial for achieving accurate prediction. Xing et al. used the bat algorithm to optimize SVM parameters, showing that this model is more accurate compared to ANN. Dai et al. [21] employed an improved particle swarm optimization algorithm to optimize the regularization parameter $C$ and standardization parameter $\sigma$ of SVM, improving the accuracy of ship power load prediction.

Although artificial intelligence models based on machine learning can achieve ship attitude prediction, they suffer from poor adaptability, and ship motion exhibits strong nonlinearity. Consequently, in practical engineering applications, there are still issues of low prediction efficiency and accuracy. Therefore, hybrid models have become the focus of current research. Li et al. [22] proposed combining the PSR method with the DSRvSVR model and CAEFOA to establish a hybrid prediction method PSR-DSRvSVR-CAEFOA for ship attitude prediction. Compared to single prediction models, this model effectively improves the accuracy of ship motion forecasting. To further address the issue of low ship attitude prediction accuracy, researchers have turned their attention to preprocessing of raw data, simplifying prediction difficulties through processing of non-stationary time series. Among these methods, empirical mode decomposition (EMD) is commonly used, which directly analyzes the signal characteristics of raw data without presetting. Zhou et al. [23] proposed a least squares support vector regression (LSSVM) model based on empirical mode decomposition (EMD) for ship attitude prediction, conducting experiments with ship data to validate the effectiveness of the model. Duan et al. [24] introduced the AR-EMD-SVR prediction model, combining the SVR model with EMD technology using an AR model for boundary effect handling. Compared to traditional AR and SVR models, this model better handles non-stationarity in ship motion. Xia et al. [5] utilized a combination prediction model based on EMD-LSTM, achieving higher accuracy in predicting nonlinear and non-stationary ship motion attitude signals compared to traditional AR and single LSTM models.

Since hybrid models have higher prediction accuracy than single models, this paper adopts a "decomposition-prediction" hybrid model, combined with the particle swarm optimization (PSO) algorithm, to form a novel CEEMDAN-PSO-SVM unmanned boat attitude prediction model. Firstly, the Completely Empirical Mode Decomposition with Adaptive Noise (CEEMDAN) is used to decompose the unmanned boat attitude into intrinsic mode functions (IMFs) and residual components, thus transforming the original attitude data into a finite number of IMF sequences. Support Vector Machine (SVM) is then employed to train and predict each subsequence. In the construction of the "decomposition-prediction" hybrid model, the PSO algorithm is utilized to optimize the parameters of the Support Vector Machine, addressing the issue of difficult parameter selection and enhancing the accuracy of the unmanned boat attitude prediction algorithm.

## 1 Random wave and unmanned boat rolling model

To study the problem of unmanned boat attitude prediction, it is essential to establish a model for the motion attitudes of unmanned boats. This involves analyzing the wave energy spectrum and the frequency response function of unmanned boat sway motion to establish the motion model of unmanned boats in various degrees of freedom.

### 1.1 Description of waves

Random ocean waves propagating on the sea surface can be viewed as the transfer of wave energy. Based on random wave theory, statistical analysis of random ocean waves has been conducted, leading to the concept of "energy spectrum," which serves as a statistical characteristic of random ocean waves. The "energy spectrum" of waves, also known as wave energy spectrum, denoted as $S_\zeta(\omega)$ [25]. The wave energy spectrum is composed of a series of unit wave energies with different frequencies, amplitudes, and directions, representing the relative frequency distribution of random ocean wave energy. In practical terms, the wave energy spectrum $S_\zeta(\omega)$ is a semi-empirical and semi-theoretical formula derived from long-term measurements of random ocean waves by researchers. Its expression varies in different seas and at different times. Common expressions include Neumann spectrum, P-M spectrum, and ITTC dual-parameter spectrum [26,27]. Among them, the ITTC dual-parameter spectrum is a wave energy spectrum formula recommended by the State Oceanic Administration of China, and its expression is as follows:

$$S_\zeta(\omega) = \frac{0.74}{\omega^5} \exp\left(-\frac{g^2}{6.28^2 h_{1/3} \omega^2}\right) \tag{1}$$

Among them, $\omega$ is the frequency of random waves, the unit is rad/s; $h_{1/3}$ is the significant wave height, the unit is m; g is the gravity acceleration, its value is 9.8m/s$^2$.

### 1.2 Unmanned boat movement analysis

When an unmanned boat navigates on the sea surface, it generates complex motions with six degrees of freedom. Therefore, when analyzing the mathematical equations of unmanned boat motion, the unmanned boat model is simplified based on the following three assumptions:

(1) The unmanned boat is a rigid body and does not undergo elastic deformation during navigation and force application;

(2) The seawater is an ideal fluid, which is not compressible, and the effects of fluid viscosity and surface tension are not considered;

(3) The motion of the unmanned boat follows linear theory, and the superposition principle can be used for solving.

With the above assumptions, the unmanned boat can be regarded as a second-order linear system, and the frequency response function of sway motion is given by:

$$W(i\omega_e) = \frac{\chi}{-\Lambda^2 + 2i\mu\Lambda + 1} \tag{2}$$

Among them, $\Lambda$ is the ratio of the random wave frequency $\omega$ of the sea wave to the natural frequency $\omega_0$ of the unmanned ship, which is called the tuning factor; $\mu$ is the ratio of the attenuation coefficient $\gamma$ to the natural frequency $\omega_0$ of the unmanned ship, which is called the dimensionless attenuation coefficient; $\chi$ is the correction coefficient.

The unmanned boat is not only influenced by wind and waves at sea but also by heading and speed. When encountering waves, its random wave encounter wave energy spectrum expression is:

$$S_\zeta(\omega_e) = \frac{S_\zeta(\omega)}{1 + 2\omega v \cos\beta / g} \tag{3}$$

Where $\omega$ is the frequency of random ocean waves, measured in rad/s; $v$ is the speed of the unmanned boat in the sea, measured in m/s; $\beta$ is the angle between the unmanned boat's direction of travel and the direction of wave propagation, measured in degrees; g is the acceleration due to gravity, with a value of 9.8m/s$^2$; $S_\zeta(\omega_e)$ is the spectral density function of random ocean waves, selected according to Equation (1).

After obtaining the wave energy spectrum encountered by random waves, you can use equation (4) to calculate the sway energy spectrum density function $S_\alpha(\omega_e)$ for a given speed and heading:

$$S_\alpha(\omega_e) = |W(i\omega_e)|^2 S_\zeta(\omega_e) \tag{4}$$

To obtain samples of the unmanned boat's attitude motion, based on the significant wave height $h_{1/3}$ under different sea conditions, samples are taken at intervals of $\Delta\omega$ in the spectral space according to Table 1. The expressions for the motion response of the unmanned boat in each degree of freedom are as follows:

$$Y(t) = \sum_{k=1}^{n} \sqrt{2S_y(\omega_k)\Delta\omega} \cos(\omega_k t + \varepsilon_k) \tag{5}$$

Where $S_y(\omega_k)$ is the energy spectrum frequency function in a certain direction; $\omega_k$ is the frequency of the $k$-th encountered wave oscillation; $\varepsilon_k$ is the phase of unmanned boat motion, and the initial phase angle $\varepsilon_k$ of each harmonic is a uniformly distributed random variable between 0 and $2\pi$.

Table 1 Wave spectrum sampling frequency band table

| Meaningful wave height $h_{1/3}$ (m) | wind speed $v$ (m/s) | Simulation band $\omega$ (rad/s) | sampling interval $\Delta\omega$ (rad/s) |
|---|---|---|---|
| $h_{1/3} \leq 2.5$ | v≤8 | 0.3<$\omega$≤3.0 | 0.1 |
| 2.5<$h_{1/3}$≤5.0 | 8<v≤12 | 0.25<$\omega$≤2.5 | 0.08 |
| $h_{1/3}$>5.0 | v>12 | 0.1<$\omega$≤1.7 | 0.06 |

In the motion of the unmanned boat's six degrees of freedom, the motion in the sway, pitch, and heave directions has the greatest impact on the stable operation of most unmanned boat-mounted instruments. Therefore, this paper takes sway motion as an example to analyze the motion response of the unmanned boat under sea state 6. The relevant parameters of the unmanned boat are shown in Table 2.

Table 2 Unmanned boat parameters

| captain(m) | Ship's width (m) | Ship speed (m/s) | sea state | wind speed(m/s) | Heading (°) |
|---|---|---|---|---|---|
| 7.5 | 2.66 | 15 | 6 | 10 | 0 |

After consulting relevant literature, the inherent frequency $\omega_0 = 1.57 \text{rad}/s$, dimensionless damping coefficient $\mu = 0.06$, and correction coefficient $\chi = 0.4$ of the sway motion of an unmanned boat navigating in sea state 6 are determined. According to Equation (5), the analytical expression for the response of the unmanned boat's sway motion is as follows:

$$R_x(t) = \sum_{k=1}^{n} \sqrt{2S_\varphi(\omega_k)\Delta\omega} \cos(\omega_k t + \varepsilon_k) \tag{6}$$

Where $S_\varphi(\omega_k)$ is the sway energy spectrum frequency function; $\omega_k$ is the frequency of the $k$-th encountered wave oscillation; $\Delta\omega$ is the sampling interval of the sway energy spectrum frequency function $S_\varphi(\omega_k)$, $\Delta\omega = 0.08$; $\varepsilon_k$ is the initial phase of the $k$-th encountered wave.

The sway vibration curve of the unmanned boat within 100 seconds under sea state 6 is shown in Figure 1.

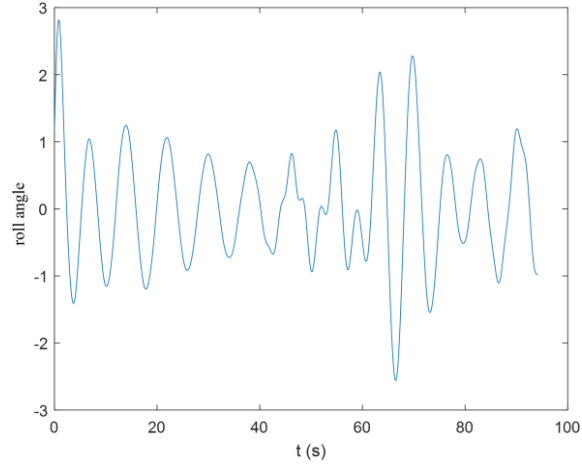

Figure 1   Unmanned boat rolling waveform

## 2 Description of unmanned boat motion attitude prediction method

### 2.1 Complete ensemble empirical mode decomposition model

Complete Ensemble Empirical Mode Decomposition with Adaptive Noise (CEEMDAN) is a signal processing method that decomposes nonlinear and non-stationary data into multiple more stable and stationary intrinsic mode function (IMF) sequences. CEEMDAN is based on the Empirical Mode Decomposition (EMD) algorithm. Empirical Mode Decomposition (EMD) is an analysis method for nonlinear and non-stationary time series proposed by Huang et al [28]. However, this method has the problem of mode mixing. To address this issue, Wu et al. [29] introduced Gaussian white noise into the signal to be decomposed, proposing Ensemble Empirical Mode Decomposition (EEMD), but this method has a large computational cost. Subsequently, Yeh et al. [30] linearly superimposed a pair of Gaussian white noises with the same amplitude but opposite signs onto the signal to be decomposed, proposing Complementary Ensemble Empirical Mode Decomposition (CEEMD). However, both of these methods leave some Gaussian white noise in the obtained Intrinsic Mode Function (IMF). To solve this problem, Torres et al. [31] proposed Complete EEMD with Adaptive Noise (CEEMDAN), which can adaptively add white noise to the signal to be decomposed, effectively addressing both the issue of mode mixing and the problem of residual auxiliary noise in the intrinsic mode components.

The motion attitude of unmanned boats exhibits nonlinearity and non-stationarity characteristics. Decomposition techniques can break it down into a series of relatively stable sequences, enabling stabilization and reducing the difficulty of predicting unmanned boat motion attitude. The specific implementation steps of CEEMDAN are as follows:

(1)   Add Gaussian white noise $v^i(t)$ to the signal to be decomposed $x(t)$ to form a new signal $x^i(t)$.

$$x^i(t) = x(t) + \omega_0 v^i(t) \qquad (7)$$

Where $\omega_0$ is the standard deviation of the white noise, and $i=1,2,\cdots\cdots,I$.

(2) Use Empirical Mode Decomposition (EMD) to decompose each component $C_1^i(t)$, and then take the average to obtain the first intrinsic mode component $C_1(t)$ and the first residual component $r_1(t)$.

$$C_1(t) = \frac{1}{I}\sum_{i=1}^{I} C_1^i(t) \qquad (8)$$

$$r_1(t) = x(t) - C_1(t) \qquad (9)$$

(3) Add Gaussian white noise $\omega_1 E_1(v^i(t))$ to the residual component $r_1(t)$ to form a new signal

$r_1 + \omega_1 E_1(v^i(t))$, where $\omega_1$ is the standard deviation of the white noise. Use EMD to decompose $r_1 + \omega_1 E_1(v^i(t))$ to obtain the second intrinsic mode component $C_2(t)$ and the second residual component $r_2(t)$.

$$C_2(t) = \frac{1}{I} \sum_{i=1}^{I} E_1\left(r_1 + \omega_1 E_1(v^i(t))\right) \tag{10}$$

$$r_2(t) = x(t) - C_2(t) \tag{11}$$

(4) Repeat the above process until the residual component becomes a monotonic function that cannot be further decomposed. The algorithm ends. At this point, \(I\) intrinsic mode components are obtained, and the original signal is decomposed into:

$$x(t) = \sum_{i=1}^{I} C_i(t) + r_I(t) \tag{12}$$

## 2.2 Basic theory of particle swarm optimization algorithm

In 1995, Kennedy et al.[32] proposed the Particle Swarm Optimization (PSO) algorithm, inspired by the life patterns of birds and fish, which is a swarm-based optimization algorithm. In PSO, each particle represents a solution, and the position of the particle in the problem space represents a feasible solution to the problem. Furthermore, there is information sharing among particles. Each particle is evaluated by a fitness function, generating corresponding fitness values. Based on these fitness values, the particles adjust their flying velocity and positions in the problem space. Through multiple iterations, the population searches for the optimal solution in the problem space. The update mechanism of particle positions is illustrated in Figure 2.

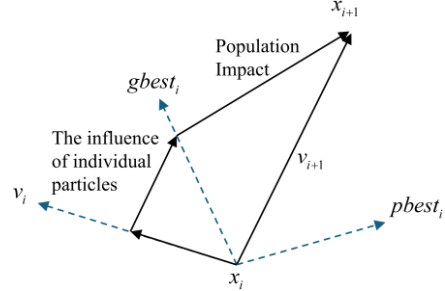

Figure 2　Update mode of particle position

Firstly, in a D-dimensional problem space, randomly generate a total of $n$ particles, where each particle's position is denoted by $x_i$ and its corresponding velocity by $v_i$, where $i=1,2,\cdots,n$. The individual best position of each particle is denoted by $pbest_i$, and the global best position of the entire population is denoted by $gbest_i$.

The formula for the position of a particle in the problem space is:

$$x_i = (x_{i1}, x_{i1}, \cdots, x_{iD}) \tag{13}$$

The formula for the velocity of a particle is:

$$v_i = (v_{i1}, v_{i1}, \cdots, v_{iD}) \tag{14}$$

The formula for updating the velocity and position of a particle is:

$$v_{i+1} = \zeta v_i + c_1 r_1 \times (pbest_i - x_i) + c_2 r_2 \times (gbest_i - x_i) \tag{15}$$

$$x_{i+1} = x_i + v_{i+1} \tag{16}$$

Where $\zeta$ is the inertia weight coefficient, $c_1$ and $c_2$ are the learning factors, and $r_1$ and $r_2$ are random numbers between 0 and 1.

2.2 Basic theory of support vector machines

In 1995, Vapnik et al. [33] introduced the Support Vector Machine (SVM) theory, which has advantages such as good generalization ability and resistance to overfitting. Initially, SVM was applied in the field of pattern recognition, mainly for classification problems. With the introduction of loss functions, SVM can also be used to solve regression problems. SVM is based on the Vapnik-Chervonenkis (VC) dimension theory and the principle of structural risk minimization (SRM), striking a balance between the complexity of the prediction model and the learning ability based on the attitude information of unmanned boats.

Given a set of training samples $D=\{(x_1,y_1),\cdots,(x_i,y_i),\cdots,(x_l,y_l)\}$, where $x_i$ is the input signal of unmanned boat attitude samples, $x_i \in R$ and $y_i$ is the corresponding output, $y_i \in R$, and $l$ is the total number of samples. For a nonlinear regression model, using kernel function $k(x_i,x_j)=\varphi(x_i)\varphi(x_j)$ to map the nonlinear sample data to a high-dimensional feature space, transforming it into a linear regression problem:

$$f(x)=\omega^T\varphi(x)+b \qquad (17)$$

Where $\omega$ is the weight vector, $\varphi(x)$ is the mapping function, and $b$ is the bias.

Under the principle of structural risk minimization, the regression problem can be mathematically represented by a constrained optimization problem. The optimization function and constraints are as follows:

$$\min \frac{1}{2}\|\omega\|^2 + C\sum_{i=1}^{l}(\xi_i^* + \xi_i)$$
$$s.t.\begin{cases} y_i - \omega\varphi(x_i) - b \leq \varepsilon + \xi_i \\ \omega\varphi(x_i) + b - y_i \leq \varepsilon + \xi_i^* \\ \xi_i, \xi_i^* \geq 0, i=1,2,\cdots l \end{cases} \qquad (18)$$

Where $C$ is the penalty factor, $\xi_i$ and $\xi_i^*$ are slack variables, $\varepsilon$ is the loss function.

Introducing Lagrange multipliers to the above equation, we can obtain the dual problem of support vector regression:

$$\max\left[-\frac{1}{2}\sum_{i=1}^{l}\sum_{j=1}^{l}(\alpha_i^* - \alpha_i)(\alpha_j^* - \alpha_j)K(x_i,x_j) - \varepsilon\sum_{i=1}^{l}(\alpha_i^* + \alpha_i) + \sum_{i=1}^{l}y_i(\alpha_i - \alpha_i^*)\right]$$
$$s.t.\begin{cases} \sum_{i=1}^{l}(\alpha_i^* - \alpha_i)=0 \\ 0 \leq \alpha_i, \alpha_i^* \leq C \end{cases} i=1,2,\cdots,l \qquad (19)$$

Where $\alpha_i^*$ and $\alpha_i$ are Lagrange multipliers, and $C$ is the penalty parameter.

At this point, substituting equation (20) into equation (16):

$$\omega = \sum_{i=1}^{l}(\alpha_i^* - \alpha_i)\varphi(x_i) \qquad (20)$$

The solution to the regression function is:

$$f(x)=\sum_{i=1}^{N}(\alpha_i^* - \alpha_i)K(x_i,x_j)+b \qquad (21)$$

Where $N$ is the number of support vectors, and $k(x_i,x_j)$ is the kernel function. The kernel function used in this paper is the most commonly used Gaussian kernel function, expressed as:

$$k(\boldsymbol{x}_i, \boldsymbol{x}_j) = \exp\left(-\frac{\|\boldsymbol{x}_i - \boldsymbol{x}_j\|^2}{2\sigma^2}\right) \tag{22}$$

Therefore, the values of the Gaussian kernel function parameter $\sigma$ and the penalty factor $C$ have a significant impact on the prediction accuracy of the support vector machine (SVM). When applying the SVM algorithm to predict the posture of unmanned boats, the first problem to address is the selection of appropriate values for the Gaussian kernel function parameter $\sigma$ and the penalty factor $C$. The Gaussian kernel function parameter $\sigma$ essentially determines the complexity of the data sample space mapping, and its value is usually within the range of $[2^{-15}, 2^3]$. The size of the penalty factor $C$ can determine the degree of data deviation and adjust the proportion of empirical risk and confidence. In general, as the penalty factor $C$ increases, the penalty imposed on the objective loss function intensifies, and the slack variable $\xi_i$ decreases, making it more prone to overfitting. However, reducing the value of $C$ may lead to excessive simplification of the model, resulting in underfitting phenomena. Based on past experience, the parameter $C$ is typically in the range of $[2^{-5}, 2^{15}]$.

## 2.3 Support vector machine algorithm based on particle swarm optimization

When using support vector machines for unmanned boat posture prediction, the accuracy is closely related to the values of the Gaussian kernel function parameter $\sigma$ and the penalty factor $C$. To find the optimal combination of support vector machine parameters and improve the accuracy of unmanned boat posture prediction, the particle swarm optimization algorithm is introduced to optimize the parameters $\sigma$ and $C$ of the prediction model. Figure 3 shows the flowchart of optimizing the support vector machine parameters based on the particle swarm optimization algorithm, and Table 3 presents the initial parameter settings for the particle swarm optimization algorithm.

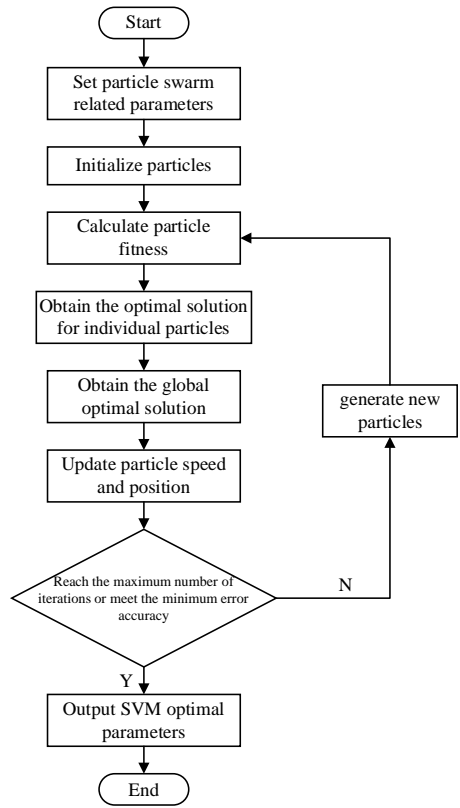

Figure 3　PSO optimization SVM flow chart

Table 3 Initialization of Particle Swarm Optimization Parameters

| Parameter Names | Values |
|---|---|
| Number of Particles | 10 |
| Learning Factor $c_1$ | 1.5 |
| Learning Factor $c_2$ | 1.7 |
| Maximum Number of Iterations | 200 |
| Inertia Weight Coefficient | 0.8 |
| Velocity Range | [-10, 10] |

To select appropriate values for the Gaussian kernel function parameter $\sigma$ and the penalty factor $C$, this paper introduces K-fold cross-validation. The basic idea of K-fold cross-validation is to divide the sample set into k equal parts and reserve one subset for testing while using the remaining k-1 subsets as training sets. This process is repeated k times, with each subset used once as the test set. The average accuracy of each round of testing results is taken as the evaluation criterion, allowing the model to adapt more widely to other data. In this paper, 3-fold cross-validation is adopted, and the reciprocal of the average accuracy of each round of testing results is used as the fitness function for the particle swarm algorithm. The fitness curve of the particle swarm algorithm is shown in Figure 4, reaching a peak of 89.2% in the 170th iteration.

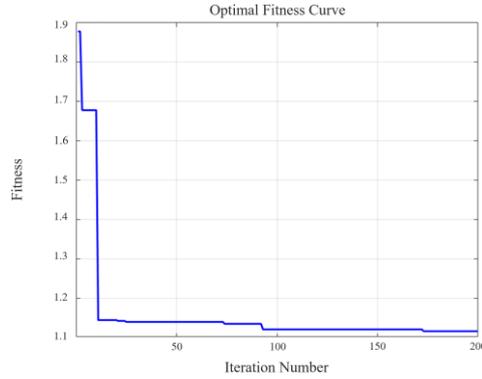

Figure 4 Particle swarm algorithm fitness curve

# 3 Design and evaluation index of unmanned boat motion attitude prediction model

## 3.1 Unmanned boat motion attitude prediction model based on CEEMDAN-PSO-SVM

Leveraging the strengths of CEEMDAN and SVM, a hybrid prediction model based on CEEMDAN-PSO-SVM for unmanned boat motion attitude prediction is designed using the "decomposition-prediction" approach.

The time series of unmanned boat motion attitude is denoted as $x(t)$, where $t=1,2,\cdots,m$. The process of unmanned boat motion attitude prediction using the CEEMDAN-PSO-SVM method, divided into training and prediction sets, is illustrated in Figure 5.

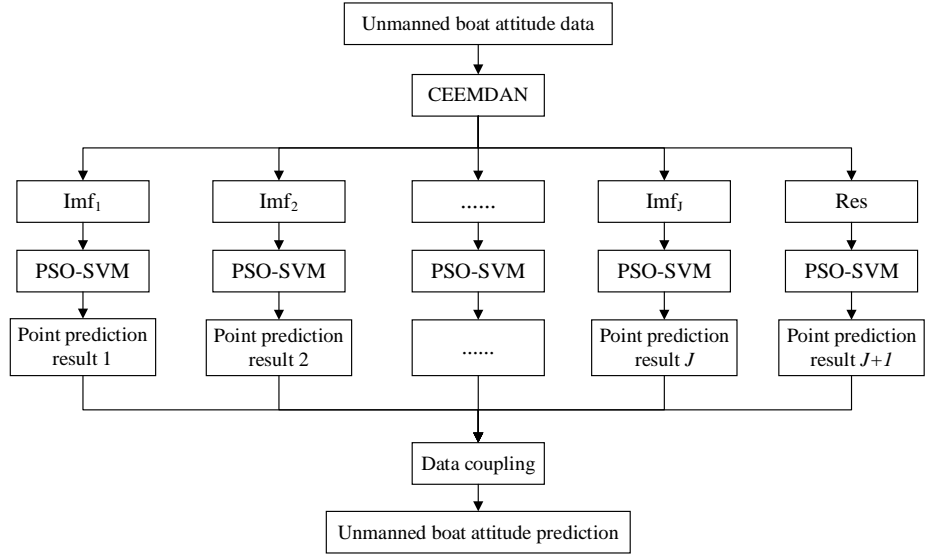

Figure 5 CEEMDAN-PSO-SVM Prediction Workflow

The specific steps are as follows:

(1) Data preprocessing: Utilize CEEMDAN to decompose $x(t)$ into I intrinsic mode functions $C_i(t)$ and one residual component $r_I(t)$. Then, divide all unmanned boat motion attitude data into training and prediction sets and normalize the data. The normalization formula is as follows:

$$x'(t) = \frac{x(t) - x(t)_{min}}{x(t)_{max} - x(t)_{min}} \tag{23}$$

Where $x'(t)$ represents the normalized data sequence, $x(t)_{max}$ denotes the maximum value in the unmanned boat motion attitude data, and \( B \) is the minimum value in the unmanned boat motion attitude data.

(2) Train the SVM model. Initialize the basic parameters of the support vector machine and particle swarm optimization algorithm. Utilize the particle swarm optimization algorithm to optimize the parameters of the support vector machine prediction model, establish the optimal support vector machine prediction model, and predict each intrinsic mode component.

(3) Data coupling. Sum up the predicted results of each intrinsic mode component to obtain the predicted results of the unmanned boat motion attitude.

## 3.2 Predictive model evaluation performance indicators

To comprehensively evaluate the forecasting performance of the wind speed prediction model proposed in this paper, the following evaluation metrics are computed for the CEEMDAN-PSO-SVM prediction results: Mean Absolute Error (MAE), Mean Absolute Percentage Error (MAPE), Mean Squared Error (MSE), Symmetric Mean Absolute Percentage Error (SMAPE), and Root Mean Square Error (RMSE). The definitions of these evaluation metrics are as follows:

$$MAE = \frac{1}{n}\sum_{i=1}^{n}\left|(y_i - \hat{y}_i)\right| \tag{24}$$

$$MAPE = \frac{100\%}{n}\sum_{i=1}^{n}\left|\frac{y_i - \hat{y}_i}{y_i}\right| \tag{25}$$

$$MSE = \frac{1}{n}\sum_{i=1}^{n}\left(y_i - \hat{y}_i\right)^2 \tag{26}$$

$$RMSE = \sqrt{\frac{1}{n}\sum_{i=1}^{n}(y_i - \hat{y}_i)^2} \qquad (27)$$

$$SMAPE = \frac{100\%}{n}\sum_{i=1}^{n}\frac{|(\hat{y}_i - y_i)|}{(|\hat{y}_i| + |y_i|)/2} \qquad (28)$$

Among them, $y_i$ is the true value of the corresponding sequence; $\hat{y}_i$ is the prediction result; $n$ is the data length of the prediction sequence. The smaller the values of MAE, MAPE, MSE, RMSE, and SMAPE indicate, the better the performance of point prediction.

# 4 Experimental results

In this study, the lateral sway of unmanned boats is selected as an example. The motion attitude sequence of unmanned boats comes from the simulated data values in Section 1.2. There are a total of 943 sample values in this dataset, with 661 samples used as training samples for the CEEMDAN-PSO-SVM combined prediction model and 282 samples used as prediction samples. First, the motion attitude sequence of unmanned boats is preprocessed according to the CEEMDAN decomposition steps to obtain 6 intrinsic mode components and 1 residual component as shown in Figure 6.

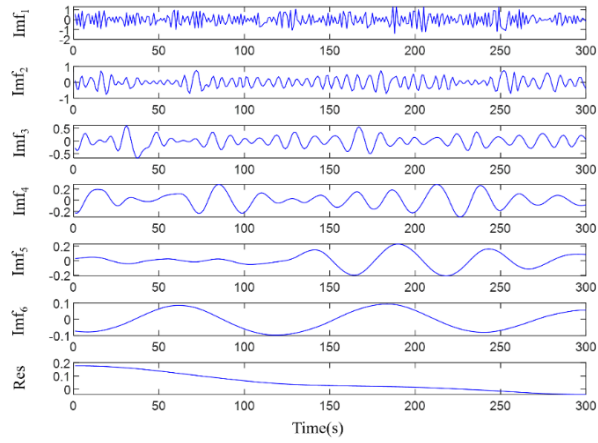

Figure 6    Rolling CEEMDAN decomposition

For each intrinsic mode component and residual component, the PSO-SVM combined model is trained separately. Subsequently, the trained models are used to predict the motion attitude of unmanned boats, resulting in the predicted lateral sway of unmanned boats as shown in Figure 7.

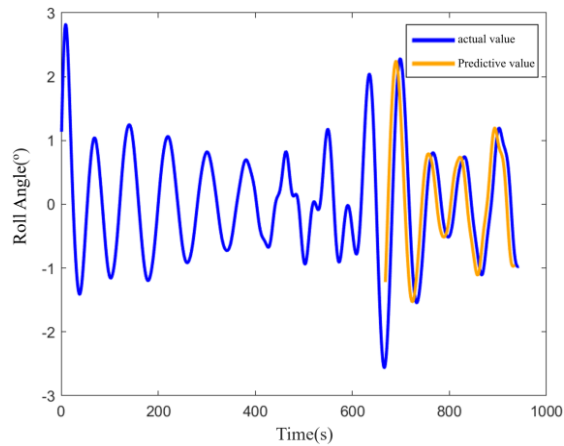

Figure 7    CEEMDAN-PSO-SVM combined model prediction results of roll

To validate the superiority of the CEEMDAN-PSO-SVM model proposed in this paper in predicting the attitude of unmanned boats, the EMD-PSO-SVM model and the CEEMDAN-SVM model are used separately to predict the lateral sway direction of unmanned boats. The predicted results are shown in Figures 8 and 9.

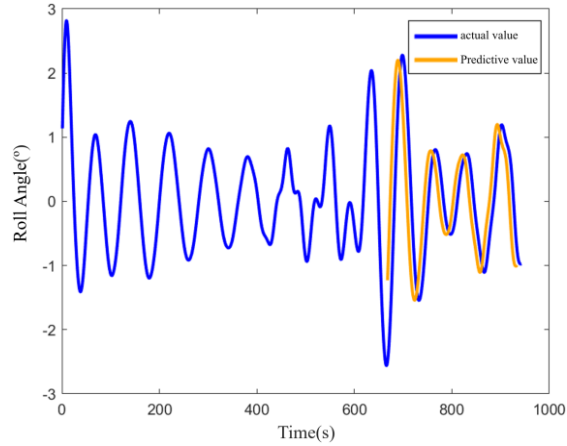

Figure 8   EMD-PSO-SVM combined model prediction results of roll

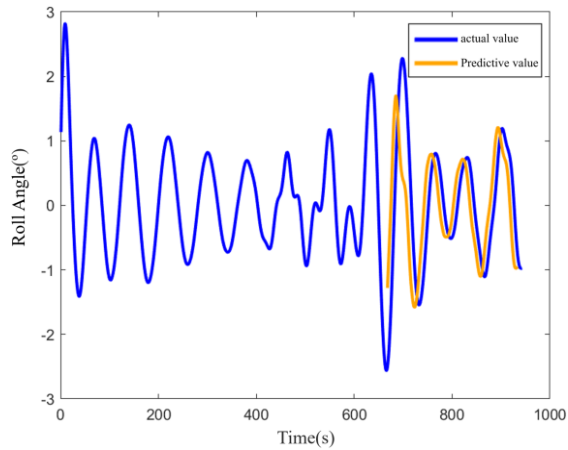

Figure 9   CEEMDAN-SVM combined model prediction results of roll

Evaluate the performance indicators of the prediction model using the evaluation criteria in Section 3.2. The evaluation results are shown in Table 4, with the best-performing sections highlighted in bold.

Table 4 Comparison of evaluation indicators of roll prediction results

|  | CEEMDAN-PSO-SVM | EMD-PSO-SVM | CEEMDAN-SVM |
|---|---|---|---|
| MAE | **0.0154** | 0.0187 | 0.0273 |
| MAPE | **0.0544** | 0.0556 | 0.0943 |
| MSE | **0.00041** | 0.00067 | 0.0015 |
| RMSE | **0.0201** | 0.0258 | 0.0347 |
| SMAPE | **0.0549** | 0.0567 | 0.0899 |

From the table data, it can be seen that the CEEMDAN-PSO-SVM model proposed in this paper exhibits good predictive performance in all five indicators of predicting the lateral sway of unmanned boats compared to the other two models. Taking the mean absolute error as an example, the predictive model proposed in this paper has improved the prediction results of unmanned boat sway by 17% and 43% compared to the results of the other two models, respectively. These experimental results validate the effectiveness of the CEEMDAN-PSO-SVM model in predicting the motion attitude of unmanned boats.

## 5 Conclusion

The characteristics of complex and unstable motion attitudes of unmanned boats result in low precision in stable platform compensation. In order to improve the precision of stable platform control and promote stable and efficient maritime operations, this paper proposes a hybrid prediction model based on the "decomposition-prediction" approach from the perspective of predicting the motion attitude of unmanned boats. Firstly, the original motion attitude data is processed using CEEMDAN, and the decomposed data consists of multiple intrinsic mode components with obvious frequency characteristics and one residual component. Then, the processed intrinsic mode components are input into SVM models for prediction separately, and PSO is used to optimize SVM to address the problem of difficult parameter selection. Finally, the coupled intrinsic mode components are combined to obtain the final prediction result. Experimental results show that compared with EMD-PSO-SVM and CEEMDAN-SVM, for example, in terms of mean absolute error, the prediction results based on the CEEMDAN-PSO-SVM combined prediction model have been improved by 17% and 43% respectively. Therefore, the hybrid prediction model adopted in this paper has better prediction performance and higher prediction accuracy.